\documentclass[manuscript,screen,publish]{acmart}

\AtBeginDocument{%
  \providecommand\BibTeX{{%
    \normalfont B\kern-0.5em{\scshape i\kern-0.25em b}\kern-0.8em\TeX}}}

\usepackage{multirow}
\usepackage{subfig}
\usepackage{color}

\begin{document}

\title{One-bit Supervision for Image Classification: Problem, Solution, and Beyond}

\author{Hengtong Hu}
\email{huhengtong.hfut@gmail.com}
\orcid{0000-0002-3095-6009}
\affiliation{%
  \institution{Hefei University of Technology}
  \city{Hefei}
  \state{Anhui}
  \country{China}
  \postcode{230000}
}
\author{Lingxi Xie}
\email{198808xc@gmail.com}
\orcid{0000-0003-4831-9451}
\affiliation{%
  \institution{Huawei Inc.}
  \city{Shenzhen}
  \state{Guangdong}
  \country{China}
}

\author{Xinyue Huo}
\email{xinyueh@mail.ustc.edu.cn}
\orcid{0000-0003-1724-9438}
\affiliation{%
  \institution{University of Science and Technology of China}
  \city{Hefei}
  \country{China}}

\author{Richang Hong}
\orcid{0000-0001-5461-3986}
\authornote{The corresponding author.}
\email{hongrc@hfut.edu.cn}
\affiliation{%
  \institution{Hefei University of Technology}
  \city{Hefei}
  \country{China}
}

\author{Qi Tian}
\orcid{0000-0002-7252-5047}
\email{tian.qi1@huawei.com}
\affiliation{%
 \institution{Huawei Inc.}
 \city{Shenzhen}
 \state{Guangdong}
 \country{China}
 }

\renewcommand{\shortauthors}{Hu and Xie, et al.}

\begin{abstract}
  This paper presents one-bit supervision, a novel setting of learning with fewer labels, for image classification. Instead of training model using the accurate label of each sample, our setting requires the model to interact with the system by predicting the class label of each sample and learn from the answer whether the guess is correct, which provides one bit (yes or no) of information. An intriguing property of the setting is that the burden of annotation largely alleviates in comparison to offering the accurate label. There are two keys to one-bit supervision, which are (i) improving the guess accuracy and (ii) making good use of the incorrect guesses. To achieve these goals, we propose a multi-stage training paradigm and incorporate negative label suppression into an off-the-shelf semi-supervised learning algorithm. Theoretical analysis shows that one-bit annotation is more efficient than full-bit annotation in most cases and gives the conditions of combining our approach with active learning. Inspired by this, we further integrate the one-bit supervision framework into the self-supervised learning algorithm which yields an even more efficient training schedule. Different from training from scratch, when self-supervised learning is used for initialization, both hard example mining and class balance are verified effective in boosting the learning performance. However, these two frameworks still need full-bit labels in the initial stage. To cast off this burden, we utilize unsupervised domain adaptation to train the initial model and conduct pure one-bit annotations on the target dataset. In multiple benchmarks, the learning efficiency of the proposed approach surpasses that using full-bit, semi-supervised supervision. 
\end{abstract}

\ccsdesc[500]{Computing methodologies~Object recognition}

\setcopyright{acmlicensed}
\acmJournal{TOMM}
\acmYear{2023} 
\acmVolume{1} 
\acmNumber{1} 
\acmArticle{1} 
\acmMonth{1} 
\acmDOI{10.1145/3633779}


\keywords{One-bit supervision, semi-supervised learning, active learning, self-supervised learning, unsupervised domain adaptation. }


\maketitle

\section{Introduction}
With the development of deep learning~\cite{lecun2015deep} in recent years, training deep neural networks has become the main methodology in the computer vision field. The urge to large-scale labeled datasets hinders their applications in the real world since collecting annotations for training data is very expensive. In particular, It is even difficult for humans to memorize all categories~\cite{russakovsky2015imagenet,zhu2017object} when the dataset contains a large number of object categories (\textit{e.g.}, ImageNet~\cite{deng2009imagenet}). In these scenarios, if the worker is asked whether an image belongs to a specified class, instead of finding the accurate class label from a large number of candidates, the annotation job would be much easier.

This paper investigates this setting which is called one-bit supervision, for one bit of information is provided by the labeler via answering the yes-or-no question. In comparison, $\log_2C$ bits of information are provided by annotating an accurate label, where $C$ is the number of classes, though we point out that the actual cost of accurate annotation is often much higher than $\log_2C\times$ that of one-bit annotation. One-bit supervision is a new challenge of learning from incomplete annotation. We expect its learning efficiency on accuracy to be superior to that of semi-supervised learning under the same amount of supervision bits. For example, in a dataset with $100$ classes, we can use ${10\mathrm{K}\times\log_2100}={66.4\mathrm{K}}$ bits of information by accurately annotating $10\mathrm{K}$ samples, or use $33.2\mathrm{K}$ bits of information by accurately annotating $5\mathrm{K}$ samples, and use the remainder $33.2\mathrm{K}$ by answering yes-or-no questions. To verify the superiority of one-bit annotation, we asked three labelers to estimate the label correctness for $100$ images ($50$ correctly labeled and $50$ wrongly labeled) from ImageNet. The average annotation time is $2.72$ seconds per image (with a precision of $92.3\%$). The average time for a full-bit annotation is around $1$ minute according to~\cite{russakovsky2015imagenet}. This validates our motivation in a many-class dataset. 

Since supervision mostly comes from guessing the label for each sample, one-bit supervision has higher uncertainty compared to the conventional setting. If a guess is correct, an accurate label of the sample is obtained, if not, only one class is eliminated from the possibilities. To efficiently learn from this setting, there are two keys that should be ensured: (i) trying to improve the accuracy of each guess to obtain more positive labels, and (ii) making full use of the failed guesses so as to not waste the negative labels. This inspires us to develop a multi-stage training framework. It makes use of off-the-shelf semi-supervised approaches to train a reasonable initial model on a small number of full-labeled data. In each of the following stages, we use up part of the supervision quota by querying with the predictions on a subset of unlabeled images selected by the 
developed sampling strategies. We add correct guesses to a set of full-labeled samples and add wrong guesses to a set of negative labels. To learn from the latter, we force the semi-supervised algorithm to predict a very low probability on the eliminated class. The model is strengthened after each stage and thus is expected to achieve higher guess accuracy in the next stage. Hence, the information obtained by one-bit supervision is significantly enriched. 

Theoretically, mining hard examples for one-bit supervision can be effective when the initial model is strong enough. Inspired by the current success of self-supervised learning in computer vision, we incorporate unsupervised pre-training with our approach to strengthen the model in each stage. 

In the training process, we fine-tune the model using pre-trained weights at every stage. To learn from the negative labels (incorrect guesses), differently, we consider it as a one-vs-rest classification problem and use a binary cross-entropy loss to optimize. This new framework makes it feasible to combine with active learning to achieve better performance. 
Nevertheless, incorporating self-supervised learning still cannot cast off the need for full-bit labels in the initial stage. In order to design a more elegant framework that conducts pure one-bit annotation on the target dataset, we utilize unsupervised domain adaptation (UDA) to obtain the initial model. By using supervision from the source domain, we can train an initial model for the target domain without any labels. Then we conduct a similar training process to obtain the final model. 
  
We evaluate our setting and approach on three image classification benchmarks, namely, CIFAR100, Mini-ImageNet, and ImageNet. For the basic framework (without unsupervised pre-training), we choose the mean-teacher model~\cite{tarvainen2017mean} as a semi-supervised baseline as well as the method used for each training stage. The results on all three datasets verify that one-bit supervision is superior to semi-supervised learning under the same bits of supervision. Additionally, with diagnostic experiments, we verify that the benefits come from a more efficient way of utilizing the information of weak supervision. For the framework with unsupervised pre-training, we conduct experiments by fine-tuning a model on ImageNet. The obvious improvement shows its effectiveness which can benefit from active learning approaches. For the framework with no need for initial full-bit labels, we evaluate it on DomainNet~\cite{peng2019moment}, the largest multi-domain dataset. The results reveal that it uses few annotations to achieve comparable performance to full-supervised training.  

A preliminary version of this manuscript appeared as~\cite{hu2020one}. The major extension of this paper is three-fold. \textbf{First, we develop a new framework that combines SSL with one-bit supervision and the experiments on ImageNet verify the boost of efficiency. Also, two strategies for class balancing are proposed, and the benefit is verified with and without self-supervised pre-training. Second, we utilize UDA to design a framework to train without initial full-bit labels and conduct only one-bit annotations on the target set. Third, we provide a mathematical foundation for our approach. Inspired by this, a strategy of mining hard examples is proposed to improve the framework beyond self-supervised pre-training}. 

The remainder of this paper is organized as follows. In Section~\ref{related work}, some related work is reviewed. In Section~\ref{approach}, the proposed basic method is introduced in detail. Section~\ref{approach:mathematical} explains the mathematical foundations of our approach. 
Section~\ref{experiment} shows the experiments. Finally, the conclusions are drawn in Section~\ref{conclusions}. 

\section{Related Work}
\label{related work}

\subsection{Semi-Supervised Learning}

Semi-supervised learning can be categorized into two types. The first one~\cite{rasmus2015semi,yu2019tangent} focuses on consistency (\textit{e.g.}, the prediction on multiple views of the same training sample should be the same) and uses it as an unsupervised loss term to guide model training. For example, $\Pi$-model~\cite{laine2016temporal} maintained two corrupted paths for the calculation of unsupervised loss. Instead of using stochastic noise, the virtual adversarial training (VAT)~\cite{miyato2018virtual} algorithm utilized adversarial perturbation which can most greatly alter the output distribution to design consistency loss. 
Instead of focusing on perturbations, the Mean-Teacher~\cite{tarvainen2017mean} averaged the weights of the teacher model to provide more stable targets for the student model. Also, some works~\cite{zhang2020wcp,kuo2020featmatch,han2021explanation} attempted to construct consistency loss from other views. 

The second type~\cite{lee2013pseudo,iscen2019label,cascante2020curriculum} assigns pseudo labels to unlabeled data, which has the same effect as entropy minimization~\cite{grandvalet2005semi}. For example, Xie~\textit{et al.}~\cite{xie2020self} proposed to generate pseudo labels using a clean teacher and train the student be noised. Rizve~\textit{et al.}~\cite{rizve2021defense} proposed UPS which leverages the prediction uncertainty to guide pseudo-label selection. There are also works combining the advantages of these two kinds of methods. MixMatch~\cite{berthelot2019mixmatch} introduced a single loss to seamlessly reduce the entropy while maintaining consistency. 
FixMatch~\cite{sohn2020fixmatch} used the pseudo labels predicted on weakly-augmented images to guide the learning on strongly-augmented images iteratively. One-bit supervision is an extension to semi-supervised learning which allows to exchange quota between fully-supervised and weakly-supervised samples, and we show that it can be more efficient.

\subsection{Active Learning}

Active learning~\cite{luo2013latent,yoo2019learning} can be roughly categorized into three types via the criterion of informative examples selection. The uncertainty-based approaches utilize a designed measurement to select samples which can decrease the model uncertainty. These measurements include the predicted probability~\cite{lewis1994sequential} and the entropy of the class posterior~\cite{wang2016cost}. The diversity-based approaches~\cite{sener2017active} select diversified data points that represent the whole distribution of the unlabeled pool. Shi~\textit{et al.}~\cite{shi2019integrating} proposed to identify a small number of data samples that best represent the overall data space by joining a sparse Bayesian model and a maximum margin machine. The expected model change~\cite{freytag2014selecting,pinsler2019bayesian,gal2017deep} approaches select samples that would cause the greatest change to the current model parameters. BALD~\cite{houlsby2011bayesian} chose data points that are expected to maximize the mutual information between predictions and model posterior. BatchBALD~\cite{kirsch2019batchbald} selected informative samples by utilizing a tractable approximation to the mutual information between a batch of samples and model parameters. One-bit supervision can be viewed as a novel type of active learning that only queries the most informative part at the class level. 

\subsection{Self-Supervised Learning}

Self-supervised learning aims to explore the intrinsic distribution of data samples via constructing a series of pretext tasks. In the early stage, researchers design some handcrafted tasks to extract features including predicting the consistency from different spatial transformations, like orientation~\cite{pathak2017learning}, rotation~\cite{gidaris2018unsupervised,malisiewicz2009beyond}, counting~\cite{noroozi2017representation}, jigsaw puzzle ~\cite{noroozi2016unsupervised} etc. Other pretasks~\cite{larsson2017colorization,zhang2016colorful} restore the original image information to achieve the same destination. Recently contrastive learning~\cite{he2020momentum,chen2020simple} attracts more and more attention on unsupervised learning field. The contrastive task requires the deep network to identify the features from the same image, which is based on that different views of an image should have consistent representation. Then many works attempt to improve it from the view of increasing the difficulty of identification, \textit{e.g.}, strong data augmentation operations~\cite{chen2020big}, a large negative gallery~\cite{chen2020improved}, and additional predictors~\cite{noroozi2018boosting}. Recently there have been some works combining self-supervised learning with semi-supervised learning and achieving better performance. Zhai~\textit{et al.}~\cite{zhai2019s4l} proposed self-supervised semi-supervised learning and used it to derive novel semi-supervised image classification methods. Chen~\textit{et al.}~\cite{chen2020big} proposed to apply "unsupervised pre-train followed by supervised fine-tuning" to semi-supervised learning. 

\subsection{Unsupervised Domain Adaptation}
As a part of transfer learning, unsupervised domain adaptation (UDA) aims to utilize the label information from the source domain to improve the performance on the unlabeled target domain. It can be roughly classified into three types, the first reduces the discrepancy between source and target domain, \textit{e.g.}, Zellinger~\textit{et al.}~\cite{zellinger2017central} achieved this by defining a new metric Central Moment Discrepancy (CMD) for matching distributions. Chen~\textit{et al.}~\cite{chen2020homm} proposed to perform higher-order moment matching for improving unsupervised domain adaptation. The second aligns the features from different domains by using adversarial loss, \textit{e.g.}, ADDA~\cite{tzeng2017adversarial}. Thirdly, some methods used self-supervised methods to assist domain adaptation. Sun~\textit{et al.}~\cite{sun2019unsupervised} proposed to perform auxiliary self-supervised tasks on both domains. TTT~\cite{sun2020test} utilized self-supervision to do test-time training. We design a new training framework by using an off-the-shelf UDA approach SCDA~\cite{li2021semantic} to train the initial model. 

\section{One-bit Supervision}
\label{approach}

In this section, we introduce the proposed one-bit supervision from four aspects. Firstly, we explain the problem setting and verify its effectiveness in section~\ref{approach:formulation}. Secondly, the main framework is elaborated in section~\ref{approach:framework}, \ref{approach:suppression}, including multi-stage training and negative label suppression. Thirdly, the two extended training paradigms incorporating with SSL and UDA are introduced detailly in section~\ref{beyond:self_supervised} and~\ref{beyond:transfer}. Finally, we discuss its relationship with prior works in section~\ref{approach:relationship}. 

\subsection{Problem Statement}
\label{approach:formulation}

In the setting of conventional semi-supervised learning, it often starts with a dataset of ${\mathcal{D}}={\left\{\mathbf{x}_n\right\}_{n=1}^N}$, where $\mathbf{x}_n$ is the $n$-th sample of image data and $N$ is the total number of training samples. We use $y_n^\star$ to denote the ground-truth class label of $\mathbf{x}_n$, and it is often unknown to the training algorithm in our setting. Specifically, we have a small set that contains $L$ samples with their corresponding $y_n^\star$ provided, and $L$ is often much smaller than $N$, \textit{e.g.}, researchers often use $20\%$ of labeled data on the CIFAR100 and Mini-ImageNet, and only $10\%$ of labels on the ImageNet, just as in Section~\ref{experiments:datasets}. That is to say,  $\mathcal{D}$ is partitioned into two subsets $\mathcal{D}^\mathrm{S}$ and $\mathcal{D}^\mathrm{U}$, where the superscripts respectively represent `supervised' and `unsupervised'.

The key insight of our research is that it is very challenging to assign an accurate label for an image when the number of classes is too large. The early user studies on ImageNet~\cite{russakovsky2015imagenet,zhu2017object} show that a labeler has difficulty memorizing all categories, which largely increases the burden of data annotation. However, if we ask the testee \textit{`Does the image belong to a specific class?'} rather than \textit{`What is the accurate class of the image?'}, the annotation cost will become much smaller. 

To verify that the new setting indeed improves the efficiency of annotation, we invite three labelers who are moderately familiar with the ImageNet-1K dataset~\cite{russakovsky2015imagenet} to do a test experiment. We use a trained ResNet-$50$ model to predict labels for the test set. Then randomly sampling $100$ images, $50$ correctly labeled and 50 wrongly labeled, for three labelers to judge if the prediction is correct. This aims to set a configuration that maximally approximates the scenario that a labeler can encounter in a real-world annotation process, meanwhile, by using a half-half data mix we avoid the labeler to bias towards either positive or negative samples. An average precision of $92.3\%$ is reported by three labelers and the average annotation time for each image is $2.72$ seconds. Since an experienced labeler can achieve a top-$5$ accuracy of $\sim\!95\%$, this accuracy is acceptable, yet around $1$ minute is taken by a full annotation for each image~\cite{russakovsky2015imagenet}, higher than $10\times$ of our cost (${\log_2 1000}\approx{10}$). Hence, the benefit of learning from a large-scale dataset is verified. 

The above motivates us to propose a new setting that combines semi-supervised learning and weakly-supervised learning. The dataset is partitioned into three parts, namely, ${\mathcal{D}}\!=\!{\mathcal{D}^\mathrm{S}\cup\mathcal{D}^\mathrm{O}\cup\mathcal{D}^\mathrm{U}}$. Here $\mathcal{D}^\mathrm{O}$ represents the weakly (one-bit) annotated subset of the dataset. Given an image from $\mathcal{D}^\mathrm{O}$ and a predicted label, the task of the labeler is to distinguish if the image belongs to the specific label. If the guess is correct, we obtain the positive (true) label $y_n^\star$ of the image, otherwise, we obtain a negative label of it which is denoted as $y_n^-$, and no further supervision of this image can be obtained. 

From the view of information theory, the labeler provides $1$ bit of supervision to the system by answering the yes-or-no question. In comparison, obtaining the accurate label requires $\log_2C$ average bits of supervision. Therefore, it alleviates the burden of annotating a single image, so one can obtain much more one-bit annotations than full-bit annotations at the same cost. Taking the CIFAR100 dataset as an example. For a common semi-supervised setting, it annotates $10\mathrm{K}$ out of $50\mathrm{K}$ training images that requires ${10\mathrm{K}\times\log_2100}={66.4\mathrm{K}}$ bits of supervision. Alternately, we can annotate $5\mathrm{K}$ images in full-bit as many as $33.2\mathrm{K}$ images in one-bit. One-bit supervision uses the same amount of supervision but can achieve higher accuracy. In addition, providing the accurate label of an image costs a labeler much more effort (more than $\log_2C\times$ that of making a one-bit annotation). Therefore, it is larger than $\log_2C:1$ for the `actual' ratio of supervision between full-supervision and one-bit supervision. That is to say, our approach actually receives a smaller amount of information under the same bits of supervision. 

\begin{figure*}[!t]
\centering
\includegraphics[width=0.95\linewidth]{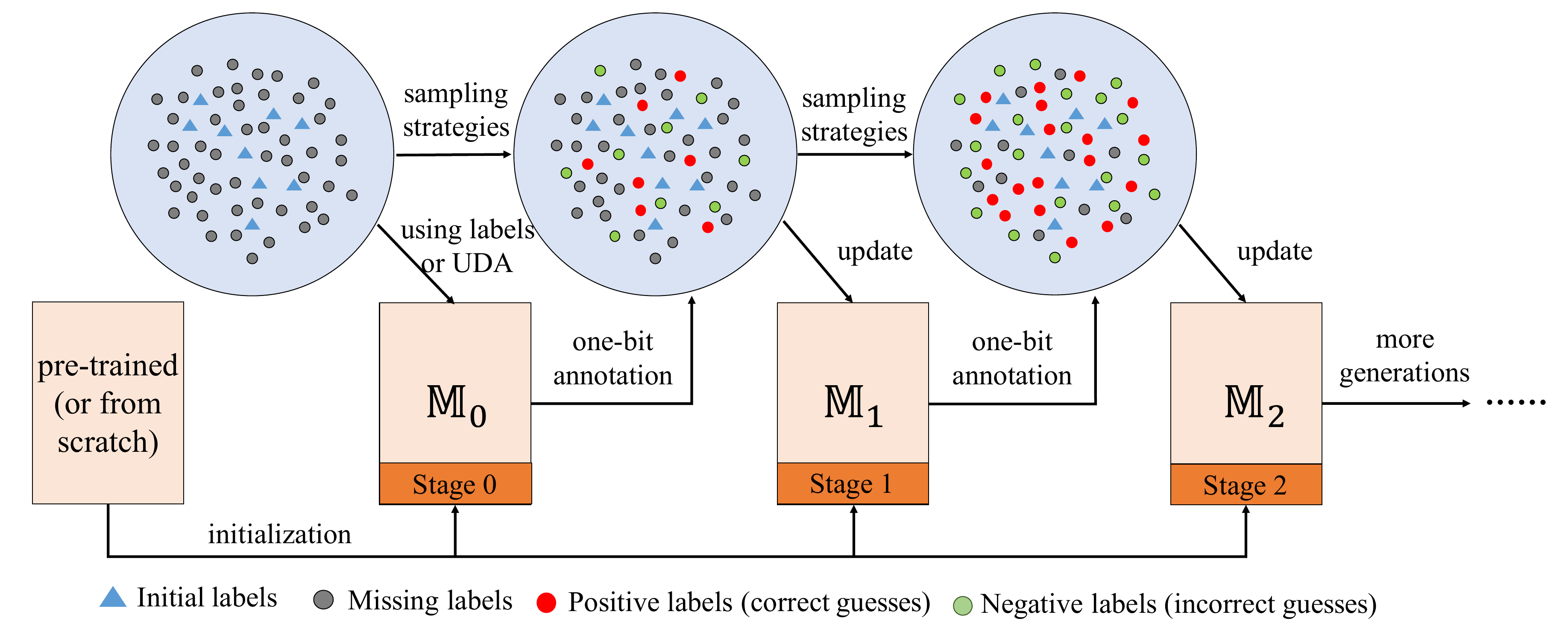}
\caption{The training procedure with one-bit supervision (with or without unsupervised pre-training, using initial full-bit labels or UDA methods). The model in each stage is initialized with the weights of a self-supervised model. At the beginning, only a small set of training samples (blue triangles) are provided with ground-truth labels and the remaining part (black circles) are unlabeled. We fine-tune the initial model on the labeled data. In each of the following stages (two stages are shown but there can be more), we send part of the unlabeled data into the current model to predict and ask the labeler to judge if the prediction is correct. Some of these samples obtain positive labels (red circles) while some obtain negative labels (green circles). This process continues until the quota of supervision is used up as scheduled. }
\label{fig:framework}
\end{figure*}

\subsection{A Multi-Stage Training Paradigm}
\label{approach:framework}

There are two important factors to one-bit supervision, namely, (i) making high-quality guesses in $\mathcal{D}^\mathrm{O}$; (ii) making use of the negative labels (incorrect guesses). The solution for the first factor is elaborated in this subsection, and the second is left in the next subsection. 

Intuitively, using more (fully or weakly) labeled training samples will improve the accuracy of a model. Considering that each image in $\mathcal{D}^\mathrm{O}$ can only be guessed once, it is straightforward to let the training procedure be partitioned into several stages. Each stage makes a prediction on a part of $\mathcal{D}^\mathrm{O}$ and then uses the results to enhance the model. This makes a generalized training algorithm as Figure~\ref{fig:framework} illustrated. We train the initial model using a semi-supervised algorithm, with $\mathcal{D}^\mathrm{S}$ as the labeled training set and $\mathcal{D}^\mathrm{O}\!\cup\!\mathcal{D}^\mathrm{U}$ as the unlabeled reference set. An off-the-shelf semi-supervised algorithm, Mean-Teacher~\cite{tarvainen2017mean}, is used to utilize the knowledge in the reference set. This makes us have a reasonable model to make predictions on $\mathcal{D}^\mathrm{O}$. 

The remaining part of training is a scheduled procedure which is composed of $T$ iterations. We define ${\mathcal{D}_{0}^\mathrm{R}}\equiv{\mathcal{D}_{0}^\mathrm{U}}$, where $\mathcal{D}_{t-1}^\mathrm{R}$ denotes the set of samples without accurate labels before $t$-th iteration. We also maintain two sets for the correct and incorrect guesses, which are denoted as $\mathcal{D}^\mathrm{O+}$ and $\mathcal{D}^\mathrm{O-}$. These two sets are initialized as $\varnothing$. In the $t$-th iteration, we first use the previous model $\mathbb{M}_{t-1}$ to make predictions on samples in $\mathcal{D}_{t-1}^\mathrm{R}$. Then, we select the subset $\mathcal{D}_t^\mathrm{O}$ from $\mathcal{D}_{t-1}^\mathrm{R}$ using the designed sampling strategies. Except for the random sampling (RS) and hard sampling (HS) strategy, we also propose a new strategy named class balance (CB) for our approach. In particular, RS selects samples from the data pool randomly, HS selects hard samples in terms of the difference of the top-$2$ prediction scores (the smaller the difference is, the harder the sample is), and CB selects by keeping a balanced number of samples in each class. RS and CB are mainly used for the experiments without unsupervised pre-training. In Section~\ref{experiment}, we will discuss the effect of these different sampling strategies. 

Then we use the predictions on $\mathcal{D}_{t}^\mathrm{O}$ to check ground-truth. We add the correctly predicted samples to $\mathcal{D}^\mathrm{O+}$ and the others to $\mathcal{D}^\mathrm{O-}$. Therefore, the entire training set is split into three parts, $\mathcal{D}^\mathrm{S}\cup\mathcal{D}^\mathrm{O+}$ (has positive labels), $\mathcal{D}^\mathrm{O-}$ (has negative labels) and $\mathcal{D}_{t}^\mathrm{U}$ (has no labels), and finally, ${\mathcal{D}_t^\mathrm{R}}={\mathcal{D}_t^\mathrm{O-}\cup\mathcal{D}_t^\mathrm{U}}$. We update $\mathbb{M}_{t-1}$ into $\mathbb{M}_t$ with the currently available supervision to obtain a stronger model. The unlabeled and fully-labeled parts of data are used as in the semi-supervised algorithm. Now we focus on utilizing the negative labels in $\mathcal{D}^\mathrm{O-}$, which will be elaborated in the following. 

\subsection{Negative Label Suppression}
\label{approach:suppression}

To make use of negative labels, we recall the Mean-Teacher algorithm~\cite{tarvainen2017mean} which maintains a teacher and a student model. It trains by making these two models produce consistent outputs, which requires no labels. Mathematically, given a training image ${\mathbf{x}}\!\in\!{\mathcal{D}}$, if ${\mathbf{x}}\!\in\!{\mathcal{D}^\mathrm{S}\!\cup\!\mathcal{D}^\mathrm{O+}}$, we compute the cross-entropy loss term first. Additionally, regardless of whether $\mathbf{x}$ has an accurate label, we compute the difference between predictions of the teacher and student model as an extra loss term. We represent the mathematical function of the student model as $\mathbf{f}\!\left(\mathbf{x};\boldsymbol{\theta}\right)$, where $\boldsymbol{\theta}$ denotes the network parameters. The teacher model is denoted by $\mathbf{f}\!\left(\mathbf{x};\boldsymbol{\theta}'\right)$ where $\boldsymbol{\theta}'$ is the moving average of $\boldsymbol{\theta}$. The loss function is written as:
\begin{equation}
\label{eqn:mean-teacher}
\begin{aligned}
& {\mathcal{L}\!\left(\boldsymbol{\theta}\right)}={\mathbb{E}_{\mathbf{x}\in\mathcal{D}^\mathrm{S}\cup\mathcal{D}^\mathrm{O+}}\ell\!\left(\mathbf{y}_n^\star,\mathbf{f}\!\left(\mathbf{x};\boldsymbol{\theta}\right)\right)} \\
& \quad\quad\quad+{\lambda\cdot\mathbb{E}_{\mathbf{x}\in\mathcal{D}}\!\left|\mathbf{f}\!\left(\mathbf{x};\boldsymbol{\theta}'\right)-\mathbf{f}\!\left(\mathbf{x};\boldsymbol{\theta}\right)\right|^2},
\end{aligned}
\end{equation}
where $\ell\!\left(\cdot,\cdot\right)$ is the cross-entropy loss and $\lambda$ is the balancing coefficient. To supplement our class balance strategy, we use a weighted cross-entropy loss to optimize, the detail is introduced in Section~\ref{beyond:self_supervised}. Here we omit the explicit notation for the individual noise added to the teacher and student model for simplicity. This means that the model's outputs on full-labeled training samples are constrained by both the cross-entropy loss and consistency loss (the idea is borrowed from knowledge distillation~\cite{hinton2015distilling,yang2018knowledge}). However, the first term is unavailable for a training sample with negative labels, so we modify $\mathbf{f}\!(\mathbf{x};\boldsymbol{\theta}')$ to inject the negative label into the second term. In particular, we let the score of the negative class be suppressed to zero. Practically, we set the \textit{logit} (before softmax) of the negative class to a large negative value to guarantee the correctness of normalization (\textit{i.e.}, the scores of all classes sum to $1$). We call this method negative label suppression (NLS). We believe that NLS is a simple and effective method that takes advantage of both the teacher model and newly added negative labels. In Section~\ref{beyond:self_supervised} we also develop a new method to utilize the negative labels. The experiments in Section~\ref{experiments:comparision} will show that NLS brings significant improvement to the one-bit supervision procedure while being naive and easy to implement. 

From a generalized view, NLS is a practical method to integrate negative (weak) labels into the framework of the semi-supervised algorithm. It is complementary to the approaches that utilize unlabeled data via a consistency loss. 


\subsection{Beyond Self-Supervised Pre-Training} 
\label{beyond:self_supervised}
In order to strengthen the initial model and mine hard examples for one-bit supervision, we propose to combine it with self-supervised learning. In particular, we integrate unsupervised pre-training into the multi-stage training framework and develop a new method to utilize negative labels. Since using MT initialized with the pre-trained weights to fine-tune makes undesirable results in experiments, we directly fine-tune a ResNet-$50$ model. As shown in Figure~\ref{fig:framework}, the training procedure is similar to that in~\ref{approach:framework}. There are two differences between them: (i) only samples with positive or negative labels are used to fine-tune, and (ii) a new loss is used to optimize the negative labels. Notably, we keep $\mathcal{D}^\mathrm{O}\!=\!\mathcal{D}_0^\mathrm{O}\!\cup\! \dots\! \cup\!\mathcal{D}_t^\mathrm{O}\!\cup\! \dots\! \cup\!\mathcal{D}_T^\mathrm{O}$ to make each image own one negative label. For the sampling strategy, we combine hard selection and class balance selection. According to their importance and predicted labels, we sort the samples in each class and select the same number of samples among classes by order. 

Similarly, the loss consists of two parts, respectively for positive and negative labels, which are written as:
\begin{equation}
\label{eqn:pre-train_loss}
\begin{aligned}
& {\mathcal{L}\!\left(\boldsymbol{\theta}\right)}={\mathbb{E}_{\mathbf{x}\in\mathcal{D}^\mathrm{S}\cup\mathcal{D}^\mathrm{O+}} w_c\cdot\ell\!\left(\mathbf{y}_n^\star,\mathbf{f}\!\left(\mathbf{x};\boldsymbol{\theta}\right)\right)} \\
& \quad\quad\quad+{\mu\cdot\mathbb{E}_{\mathbf{x}\in\mathcal{D}^\mathrm{O-}}\left(1-\mathbf{y}_n^B \right)\cdot\mathrm{log}\!\left(1-\sigma\!\left( \mathbf{f}\!\left(\mathbf{x};\boldsymbol{\theta}\right) \right)  \right)}. 
\end{aligned}
\end{equation}
The left part of Eq.~\eqref{eqn:pre-train_loss} is a weighted cross-entropy loss for positive labels. To supplement the strategy of class balance sampling, we use the weighted loss to alleviate the class imbalance issue for experiments both with and without unsupervised pre-training. The corresponding weight for class $c$ is defined as $w_c = \frac{m_c}{\mathrm{max}\left(m_0,\ldots,m_C \right)}$, where $m_c$ denotes the number of samples in $c$-th class. The right part of Eq.~\eqref{eqn:pre-train_loss} is a binary cross-entropy loss which is used to optimize negative labels. $\mu$ is the weight parameter of the negative loss, and $\mathbf{y}_n^B$ is the binary label which denotes whether $\mathbf{x}_n$ has an accurate label and $\sigma\!\left(\cdot\right)$ represents the sigmoid function. Here we don't use the formulation of negative label suppression in~\ref{approach:suppression} because the consistent loss is not maintained. It is believed that the type of negative loss should be adjusted to the baseline method. 

\subsection{Beyond Unsupervised Domain Adaptation} 
\label{beyond:transfer} 
We develop a novel framework that trains without initial full-bit labels and conducts pure one-bit annotation on the target dataset via unsupervised domain adaptation. In particular, firstly we use an off-the-shelf domain adaptation approach, \textit{e.g.}, the recently proposed SCDA~\cite{li2021semantic}, to train an initial model for target dataset $\mathcal{D}^\mathrm{T}$. Here SCDA proposed to efficiently align the feature distributions by encouraging the model to concentrate on the most principal features. Also, using source data from different domains will obtain a model with different accuracy. After obtaining the initial model $\mathbb{M}_{0}$ without using any full-bit annotation on the target dataset, \textit{i.e.}, the supervised part on target domain $\mathcal{D}^\mathrm{T-S}$ is $\varnothing$, we use it to conduct one-bit annotation on whole $\mathcal{D}^\mathrm{T}$. Similar to that in section~\ref{approach:framework}, we obtain the positive and negative labels for samples in $\mathcal{D}^\mathrm{T}$. Then we utilize all available information to update the model. Here we train the next stage model in two ways, \textit{i.e.}, using source data or not. For training using source data, we add two extra losses for the target dataset to the original loss in SCDA~\cite{li2021semantic}, which respectively are cross entropy for positive labels and binary cross-entropy for negative labels. 

For training without source data, two strategies are adopted in different stages, \textit{i.e.}, using semi-supervised training (\textit{e.g.}, FixMatch~\cite{sohn2020fixmatch}) when correctly guessed labels are less than $80\%$ of total images, or supervised fine-tuning when the number is more than $80$ percent. Each stage we conduct one-bit annotation for the samples without positive labels, \textit{i.e.}, for $\mathcal{D}_{t-1}^\mathrm{R}$ on $t$-th stage, and the training continues until we obtain satisfied accuracy. In addition, we introduce active learning into this framework, to verify the least supervision can be used to approach the full-supervised performance. Particularly, we calculate the standard deviation for the predictive probabilities of each sample after multiple different augmentations and select half of the samples with the largest deviation values to annotate in each stage. We utilize this framework to show the superiority of one-bit supervision in annotations saving when approaching upper-bound performance (supervised training using all labels).

\subsection{Relationship with Prior Work}
\label{approach:relationship}

The development of deep learning, in particular training deep neural networks~\cite{he2016deep,zagoruyko2016wide,huang2017densely}, is built upon the need for large collections of labeled data. To mitigate this problem, researchers proposed semi-supervised learning~\cite{fergus2009semi,guillaumin2010multimodal} and active learning~\cite{atlas1990training,lewis1994sequential} as effective solutions to utilize the larger amounts of unlabeled data. From the view of entropy minimization, semi-supervised learning and active learning can be considered as two ways to achieve the same goal, and some works indeed combine these two approaches into one~\cite{rizve2021defense,gao2020consistency}. 
One-bit supervision also can be regarded as one of these approaches. Also, the utilization of negative labels is related to the approaches in \cite{chen2020negative,kim2019nlnl}. However, different from those that randomly select negative labels among all classes, our approach obtains hard negative labels by annotating and brings more improvement to training. 

One-bit supervision shares a similar idea of using human verification for the model predictions with \cite{papadopoulos2016we}. Additionally, the proposed multi-stage training algorithm is related to knowledge distillation (KD)~\cite{romero2014fitnets,hinton2015distilling,hu2020creating}, which iteratively trains the same model while absorbing knowledge from the previous stage. KD was used for model compression originally, but recent years have witnessed its application for optimizing the same network across generations~\cite{furlanello2018born,yang2018knowledge}. For one-bit supervision, new supervision comes in after each stage, but the efficiency of supervision is guaranteed by the previous model, which is a generalized way of distilling knowledge from the previous model and fixing it with weak supervision. 

\section{Mathematical Foundations}
\label{approach:mathematical}

One-bit supervision can be viewed as a novel type of active learning that only queries the most informative part at the class level. Here we briefly introduce the theory of active learning and then deduce the theory foundations of one-bit supervision from three aspects. First, one-bit annotation is superior to full-bit annotation under the same cost when satisfying specific conditions. Second, the best solution to one-bit supervision is to query by using the class with the largest predicted probability. Third, class-level query and sample-level query can be combined when satisfying specific conditions. 

Given a model $\mathbb{M}$ and unlabeled data $\mathcal{D}^\mathrm{U}$, active learning aims to utilize an acquisition function $\alpha\left(\mathbf{x}, \mathbb{M}\right)$ to select samples $\mathbf{x} \in \mathcal{D}^\mathrm{U}$ to query: 
\begin{equation}
\label{eqn:active_learning}
\mathbf{x}^* = \mathrm{argmax}_{\mathbf{x} \in \mathcal{D}^\mathrm{O}} \,\alpha\left(\mathbf{x}, \mathbb{M}\right)
\end{equation}
Here we focus on the ideas that define $\alpha\left(\mathbf{x}, \mathbb{M}\right)$ based on uncertainty. For the classification task, we want to look for images with high predictive variance to label which can decrease model uncertainty. Then the acquisition function can be defined to choose samples that maximize the predictive entropy, 
\begin{equation}
\label{eqn:acquisition_function}
\begin{aligned}
\mathbb{H}&\left[\mathbf{y}\!\mid\! \mathbf{x},\mathcal{D}^\mathrm{S}  \right] \coloneqq \\
&-\sum_c p\left(y_c\!=\!1\!\mid\! \mathbf{x},\mathcal{D}^\mathrm{S}  \right) \mathrm{log}\, p\left(y_c\!=\!1\!\mid\! \mathbf{x},\mathcal{D}^\mathrm{S}  \right). 
\end{aligned}
\end{equation}

Here $\mathbf{y}\!=\![y_1,\ldots,y_{C}]$ is the one-hot vector. When $y_c\!=\!1$ it represents the sample belongs to class $c$, $c\!\in\! [1,\ldots,C]$, and $C$ is the total number of classes. The $p\left(y_c\!=\!1\!\mid\! \mathbf{x},\mathcal{D}^\mathrm{S} \right)$ denotes the probability of $y_c\!=\!1$ when given $\mathbf{x}$ and $\mathcal{D}^\mathrm{S}$. For one-bit supervision, we aim to look for the classes with high predictive variance to query. Therefore we can develop our system as: 
\begin{equation}
\label{eqn:system_1bit}
\mathrm{argmax}_{c\in [1,\ldots,C]} \,\alpha\left(\mathbf{x}, \mathbb{M}\right). 
\end{equation}
We define the acquisition function as  
\begin{equation}
\label{af_1bit}
\begin{aligned}
\mathbb{H}&\left[y_c\!\mid\! \mathbf{x},\mathcal{D}^\mathrm{S}  \right] \coloneqq \\
&-p\left(y_c\!\mid\! \mathbf{x},\mathcal{D}^\mathrm{S}  \right) \mathrm{log}\, p\left(y_c\!\mid\! \mathbf{x},\mathcal{D}^\mathrm{S}  \right) \\
&- \left(1-p\left(y_c\!\mid\! \mathbf{x},\mathcal{D}^\mathrm{S}  \right) \right)\mathrm{log}\left(1-p\left(y_c\!\mid\! \mathbf{x},\mathcal{D}^\mathrm{S}  \right) \right).
\end{aligned}
\end{equation}
For simplicity we abbreviate $p\left(y_c\!\mid\! \mathbf{x},\mathcal{D}^\mathrm{S}\right)$ as $p_c\!\in\![0,1], \sum_{c=1}^{C} p_c\!=\!1$, and denote $\mathbb{H}\left[y_c\!\mid\! \mathbf{x},\mathcal{D}^\mathrm{S}  \right]$ as $\mathbb{H}\left(p_c \right)$. Then we have 
\begin{equation}
\label{simplicity}
\mathbb{H}\left(p_c \right) = -p_c \mathrm{log}\, p_c - \left(1-p_c \right)\mathrm{log}\left(1-p_c \right). 
\end{equation}

First, we compare the efficiency of one-bit and full-bit annotation. In Section~\ref{approach:formulation} we point out that the former provides $1$ bit of information while the latter provides $\log_2C$ bits. Therefore, their average entropy production brought by each bit can be denoted by $\mathbb{H}\left[y_c\!\mid\! \mathbf{x},\mathcal{D}^\mathrm{S} \right]/1$ and $\mathbb{H}[\mathbf{y}|\mathbf{x},\mathcal{D}^S]/\log_2C$, respectively. Next, we give a theorem to define the relationship between them and then prove it. 
\begin{theorem}
\label{theo:superior}
Suppose $C\!\geqslant\! 2$, $\varphi(C)$ is a function of $C$, $\exists\, \varphi(C)\!\leqslant\!\frac{1}{2}$ such that $\forall x\!\in\!\mathcal{D}^U$, if $p_c \geqslant \varphi(C)$, 
\begin{equation}
\label{eqn:thm1}
\mathbb{H}[y_c|\mathbf{x},\mathcal{D}^S]\geqslant\frac{\mathbb{H}[\mathbf{y}|\mathbf{x},\mathcal{D}^S]}{\log_2C}
\end{equation}
\end{theorem}

\begin{proof}
Note that,  
\begin{equation}
\begin{aligned}
\mathbb{H}[\mathbf{y}|\mathbf{x},\mathcal{D}^S] &= -p_c\log{p_c}-\sum_{i\neq c}p_{i}\log{p_{i}}\\
&\leqslant -p_c\log{p_c}-(C-1)\frac{1-p_c}{C-1}\log\frac{1-p_c}{C-1}\\
&= \mathbb{H}[y_c|\mathbf{x},\mathcal{D}^S]+(1-p_c)\log(C-1). 
\end{aligned}
\end{equation}
Eq.~\eqref{eqn:thm1} holds when $\mathbb{H}(p_c)\!\geqslant\!\frac{\mathbb{H}(p_c)\!+\!(1-p_c)\log(C\!-\!1)}{\log_2C}$, which is equivalent to 
\begin{equation}
\label{eqn:equivalent}
\frac{-p_c}{1-p_c}\log p_c-\log(1-p_c) \geqslant \frac{\log(C-1)}{\log C - 1}
\end{equation}
Let $f(p)\!\triangleq\!\frac{-p}{1-p}\log p\!-\!\log(1\!-\!p)$, we have $f'(p)\!=\!\frac{-\!\log p}{(p-1)^2}\geqslant 0$ and $f(\frac{1}{2}) = 2\geqslant\frac{\log(C-1)}{\log C - 1}\!\iff\! C^2-4C+4\geqslant0$ always holds. Hence, $\exists\, \varphi(C)\leqslant\frac{1}{2}$, $\forall p_c>\varphi(C)$,
\begin{equation}
\label{eqn:verification}
\mathbb{H}(p_c)\geqslant\frac{\mathbb{H}(p_c)+(1-p_c)\log(C-1)}{\log_2C}\geqslant \frac{\mathbb{H}[\mathbf{y}|\mathbf{x},\mathcal{D}^S]}{\log_2C}
\end{equation}
\end{proof}

So far, we give a definition for when one-bit annotation brings more average entropy production, \textit{i.e.}, is more efficient, than full-bit annotation. For a dataset with $100$ classes, the numerical solution of Eq.~\eqref{eqn:equivalent} is $p_c\geqslant0.28$. Moreover, the restriction for $p_c$ will be eased when $C$ becomes larger. Hence, the condition for $p_c$ in Theorem~\eqref{theo:superior} is easy to satisfy in real-world applications, which means one-bit annotation is superior to full-bit annotation in most cases. Next, we will verify that solving Eq.~\eqref{eqn:system_1bit} is equivalent to solve 
\begin{equation}
\label{eqn:solution_1bit}
\mathrm{argmax}_c \,p\left(y_c\!\mid\! \mathbf{x},\mathcal{D}^\mathrm{S}  \right). 
\end{equation}
From the definition of $\mathbb{H}\left(p_c \right)$ we can know that it satisfies $\mathbb{H}\left(\frac{1}{2}+p_c \right)\!=\!\mathbb{H}\left(\frac{1}{2}-p_c \right)$. The derivative of $\mathbb{H}$ with respect to $p_c$ is $\frac{\partial \mathbb{H}}{\partial p_c}\!=\!\mathrm{log}\left(\frac{1}{p_c}-1 \right)$. 
Then we know $\mathbb{H}\left(p_c\right)$ increases when $p_c\!\in \![0,\frac{1}{2}]$ and decreases when $p_c\!\in\![\frac{1}{2},1]$. $p_c\!=\!\frac{1}{2}$ is the maximal point. Suppose $c'$ is the solution of Eq.~\eqref{eqn:solution_1bit}. We prove that $c'$ also maximizes Eq.~\eqref{eqn:system_1bit}. If there is a $c^*\!\neq\! c'$ such that $\mathbb{H}[p_{c^*}]>\mathbb{H}[p_{c'}]$, we have $p_{c^*}\!\in\!\left(\min\left(p_{c'},1-p_{c'}\right),\max\left(p_{c'},1-p_{c'}\right)\right)$ due to the monotonicity and symmetry of $\mathbb{H}$. On the other hand, $p_{c^*}\!<\!p_{c'}$ since $c'$ maximize $p_c$ and $p_{c'}\!\neq\! p_{c^*}$. So $p_{c^*}\!\in\!\left(1\!-\!p_{c'},p_{c'}\right)$, which deduce that $p_{c^*}\!+\!p_{c'}\!>\!1$, contradict to $\sum_cp_c\!=\!1$. Therefore, by solving Eq.~\eqref{eqn:solution_1bit} the most informative part in class-level will be chose for the sample. 

\begin{theorem}
\label{theo:combine}
$\frac{\mathbb{H}[y_c|\mathbf{x},\mathcal{D}^S]}{\mathbb{H}[\mathbf{y}|\mathbf{x},\mathcal{D}^S]}$ increases to $1$ as $p_c$ increases to $1$.  
\end{theorem}
Finally, Theorem~\eqref{theo:combine} gives a definition for when it can be effective to query at the class level for the mined hard examples. The proof of this theorem is omitted for its simplicity. Notably, $\frac{\mathbb{H}[y_c|\mathbf{x},\mathcal{D}^S]}{\mathbb{H}[\mathbf{y}|\mathbf{x},\mathcal{D}^S]}\!\rightarrow\! 1$ represents the entropy production brought by one-bit annotation increases to approach which full-bit annotations brings. That being said, annotating the hard examples in a one-bit setting can be effective when $p_c$ is large enough. More important, the certainty of the model also decides the holding of Theorem~\eqref{theo:combine}, because a weak model can make a wrong prediction with high confidence. This inspires us to increase $p_c$ by enhancing the initial model without using extra labels.

\section{Experiments}
\label{experiment}

In this section, we validate the proposed approach via the elaborate experiments. The used datasets and implementation details are introduced first. Then we compare the main framework of our approach to the full-bit and semi-supervised methods to show its superiority. Next, we analyze the performance of using the different number of training stages and guessing strategies. These two parts belong to the conference version. Followingly, we show the improvement brought by the strategies of class balance and the performance of two extended frameworks which incorporate with SSL and UDA. These are newly proposed in this manuscript. 

\subsection{Datasets and Implementation Details}
\label{experiments:datasets}

The main experiments are conducted on three popular image classification benchmarks, namely, CIFAR100, Mini-Imagenet, and Imagenet. CIFAR100~\cite{krizhevsky2009learning} contains $60\mathrm{K}$ images in which $50\mathrm{K}$ for training and $10\mathrm{K}$ for testing. All of them are $32\times 32$ RGB images and uniformly distributed over $100$ classes. Mini-ImageNet contains images from $100$ classes with resolution $84\times84$ , and the training/testing split created in~\cite{ravi2016optimization} is used, which consists of $500$ training images, and $100$ testing images per class. For ImageNet~\cite{deng2009imagenet}, the competition subset~\cite{russakovsky2015imagenet} is used which contains $1\mathrm{K}$ classes, $1.3\mathrm{M}$ training images, and $50\mathrm{K}$ testing images. We use all images of high resolution and pre-process them into $224\times224$ as network inputs. The experiments involved with UDA are conducted on DomainNet~\cite{peng2019moment}, which contains about $120\rm{,}906$ images of $345$ categories in each of the six domains, and we mainly use three of them, namely Clipart, Quickdraw, and Real. 

For experiments without unsupervised pre-training, we build our baseline by following Mean-Teacher~\cite{tarvainen2017mean}, a previous semi-supervised approach. It assumes that there is a labeled small subset, $\mathcal{D}^\mathrm{S^\prime}$. $\left|\mathcal{D}^\mathrm{S^\prime}\right|$ is $20\%$ of the training set for CIFAR100 and Mini-ImageNet, and $10\%$ for ImageNet. By allowing part of the annotation to be one-bit, we reschedule the assignment which results in two subsets, $\left|\mathcal{D}^\mathrm{S}\right|$ and $\left|\mathcal{D}^\mathrm{O}\right|$, satisfying ${\left|\mathcal{D}^\mathrm{S\prime}\right|}\approx{\left|\mathcal{D}^\mathrm{S}\right|+\left|\mathcal{D}^\mathrm{O}\right|/\log_2C}$. Table~\ref{tab:settings} shows the detailed configuration for the four datasets. For CIFAR100, we use a $26$-layer deep residual network~\cite{he2016deep} with Shake-Shake regularization~\cite{gastaldi2017shake}. For Mini-ImageNet and ImageNet, a $50$-layer residual network is used for training. For DomainNet, ResNet-$101$ is used as the backbone. The experiments are trained $180$ epochs for CIFAR100 and Mini-ImageNet and $60$ epochs for ImageNet. As in~\cite{tarvainen2017mean}, we compute the consistency loss using the mean square error in each stage. The consistency parameter is $1\rm{,}000$ for CIFAR100 and $100$ for Mini-ImageNet and ImageNet. We simply follow the original implementation for other hyper-parameters, except adjusting the batch size to fit our hardware (\textit{e.g.}, eight NVIDIA Tesla-V100 GPUs for ImageNet experiments). 

For experiments with unsupervised pre-training on ImageNet, we also use a $50$-layer residual network to fine-tune. The setting of dataset split is just as the Table~\ref{tab:settings} shows. The initial learning rate is set to $10^{-4}$ for the backbone and $1$ for the fc layer for each training stage. We train the model for $100$ epochs for each stage. The weight parameter of the negative loss is set to $0.1$. The whole batch size including both the positively and negatively labeled data is set to $1024$. For experiments on DomainNet, the setting of domain adaptation just follows SCDA~\cite{li2021semantic}, and all the experiments are trained for $100$ epochs. 

\begin{table*}[!t]
\renewcommand{\arraystretch}{1.3}
\caption{The data split for semi-supervised and one-bit-supervised approaches, where we compare the total numbers of supervision bits. The other data splits are investigated in Section~\ref{experiments:stages}.}
\label{tab:settings}
\begin{center}
\setlength{\tabcolsep}{0.3cm}
\begin{tabular}{lcccccccc}
\hline
\multirow{2}{*}{Dataset} & \multirow{2}{*}{$C$} & \multirow{2}{*}{$\log_2C$} & \multirow{2}{*}{$\left|\mathcal{D}\right|$} & \multicolumn{2}{c}{semi-supervised} & \multicolumn{3}{c}{one-bit-supervised} \\
\cline{5-9}
{} & {} & {} & {} & $\left|\mathcal{D}^\mathrm{S^\prime}\right|$ & \# of bits & $\left|\mathcal{D}^\mathrm{S}\right|$ & $\left|\mathcal{D}^\mathrm{O}\right|$ & \# of bits \\
\hline
CIFAR100 & $100$ & $6.6439$ & $50\mathrm{K}$ & $10\mathrm{K}$ & $66.4\mathrm{K}$ & $3\mathrm{K}$ & $47\mathrm{K}$ & $66.9\mathrm{K}$ \\
Mini-ImageNet & $100$ & $6.6439$ & $50\mathrm{K}$ & $10\mathrm{K}$ & $66.4\mathrm{K}$ & $3\mathrm{K}$ & $47\mathrm{K}$ & $66.9\mathrm{K}$ \\
ImageNet & $1\rm{,}000$ & $9.9658$ & $1281\mathrm{K}$ & $128\mathrm{K}$ & $1276\mathrm{K}$ & $30\mathrm{K}$ & $977\mathrm{K}$ & $1276\mathrm{K}$ \\
DomainNet & $345$  & $8.4305$  & $120\rm{,}906$  & $120\rm{,}906$  & $1019.3\mathrm{K}$  & $0$  & $120\rm{,}906$  & -  \\ 
\hline
\end{tabular}
\end{center}
\end{table*}

\subsection{Main Results Compared to Full-bit, Semi-supervised Supervision}
\label{experiments:comparision}

First, we compare our approach to Mean-Teacher~\cite{tarvainen2017mean}, a semi-supervised baseline that utilizes full-bit annotations. Results are summarized in Table~\ref{tab:comparison}. We can observe that, the one-bit supervision baseline (with two stages, without NLS) is inferior to full-bit supervision. Notably on CIFAR100, it actually obtains more accurate labels (the number is $3\mathrm{K}+25.3\mathrm{K}$, compared to $10\mathrm{K}$ used in the semi-supervised baseline), but the obtained correct guesses do not contribute much to training. This issue becomes more obvious on Mini-ImageNet and ImageNet since they have weaker initial models and more incorrectly guessed images. We attribute this to that these samples contribute little new knowledge to training because (i) the model has already learned how to classify these samples, and (ii) these samples are relatively easy compared to the incorrectly predicted ones. Therefore, it is crucial for one-bit supervision to make use of negative labels.

We next investigate negative label suppression (NLS), an approach used to extract knowledge from incorrect guesses. As shown in Table~\ref{tab:comparison}, significant improvement is brought by simply suppressing the score of the incorrect class for each element in $\mathcal{D}^{\mathrm{O}-}$. In particular, this brings $4.37\%$, $5.86\%$, and $4.76\%$ accuracy gains on CIFAR100, Mini-ImageNet, and ImageNet respectively, when compared to the two-stage baseline. It reveals that though only filtering out one out of $100$ or $1\rm{,}000$ classes, the negative labels can provide important supervision for semi-supervised learning, and the key contribution is to avoid the teacher and student models from arriving in a wrong consensus. 

\begin{table*}[!t]
\renewcommand{\arraystretch}{1.4}
\caption{Comparison of accuracy ($\%$) to our baseline, Mean-Teacher~\cite{tarvainen2017mean}, and some state-of-the-art semi-supervised learning algorithms, which corresponds to the main framework of our approach. On all datasets, we report the top-$1$ accuracy. In our multi-stage training process, we report the accuracy of the initial stage (using Mean-Teacher for semi-supervised learning) as well as after each one-bit supervision stage. The discussion of using one or two stages is in Section~\ref{experiments:stages}. NLS indicates negative label suppression (see Section~\ref{approach:suppression}). CB represents the strategies of class balance.}
\label{tab:comparison}
\begin{center}
\setlength{\tabcolsep}{0.4cm}
\begin{tabular}{lccc}
\hline
Method     & CIFAR100 & Mini-ImageNet & ImageNet \\
\hline
$\Pi$-Model~\cite{laine2016temporal} & $56.57$ (ConvNet-13) & - & - \\
DCT~\cite{qiao2018deep} & $61.23$ (ConvNet-13) & - & $53.50$ (ResNet-18) \\
LPDSSL~\cite{iscen2019label} & $64.08$ (ConvNet-13) & $42.65$ (ResNet-18) & - \\
\hline
Mean Teacher~\cite{tarvainen2017mean} & $69.76$ (ResNet-26) & $41.06$ (ResNet-50) & $58.16$ (ResNet-50) \\
Ours (1-stage base) & $51.47\rightarrow66.26$ & $22.36\rightarrow35.88$ & $47.83\rightarrow54.46$ \\
\quad +NLS          & $51.47\rightarrow71.13$ & $22.36\rightarrow38.30$ & $47.83\rightarrow58.52$ \\
Ours (2-stage base) & $51.47\rightarrow64.83\rightarrow69.39$ & $22.36\rightarrow33.97\rightarrow39.68$ & $47.83\rightarrow54.04\rightarrow55.64$ \\
\quad +NLS          & $51.47\rightarrow67.82\rightarrow73.76$ & $22.36\rightarrow37.92\rightarrow45.54$ & $47.83\rightarrow57.44\rightarrow60.40$ \\ 
\textbf{\quad +CB}          & $\textbf{51.47}\rightarrow\textbf{68.14}\rightarrow\textbf{73.89}$ & $\textbf{22.36}\rightarrow\textbf{38.17}\rightarrow\textbf{46.82}$ & $\textbf{47.83}\rightarrow\textbf{59.41}\rightarrow\textbf{65.10}$ \\ 
\hline
\end{tabular}
\end{center}
\end{table*}

In summary, our approach achieves favorable performance in one-bit supervision with two-stage training and negative label suppression. In particular, under the same bits of supervision, we achieve $4.00\%$, $4.48\%$, and $2.24\%$ accuracy gain over the semi-supervised baseline, on CIFAR100, Mini-ImageNet, and ImageNet respectively. Hence, it is cleaner for the effectiveness of our learning framework, as well as the multi-stage training algorithm. Though the experiments have only tested on top of the Mean-Teacher, we believe that this pipeline can be generalized to other semi-supervised approaches as well as other network backbones. 

\subsection{Number of Stages and Guessing Strategies}
\label{experiments:stages}

Next, we do some ablation studies for one-bit supervision, namely, the number of stages used, the size of $\mathcal{D}^\mathrm{S}$, the sampling strategy on $\mathcal{D}^\mathrm{O}$, \textit{etc}. 

\subsubsection{Analysis on Number of Training Stages}
We compare the experiments of one-stage training and two-stage training. Specifically, the former uses up the quota of one-bit annotations at one time, and the latter split the quota into two parts to first train an intermediate model and then iterate to obtain the final model. Table~\ref{tab:comparison} shows the results of three benchmarks. It is clear that the advantage of two-stage training is that more positive labels can be found by training a stronger intermediate model. For one-stage training, the numbers of correct guesses are around $23.2\mathrm{K}$, $9.8\mathrm{K}$, and $470\mathrm{K}$, respectively on CIFAR100, Mini-ImageNet, and ImageNet. While using two-stage training, these numbers become $25.3\mathrm{K}$, $12.2\mathrm{K}$, and $475\mathrm{K}$. As a result, compared with one-stage training, the two-stage training boosts the accuracy of the final model by $2.63\%$, $7.24\%$, and $3.46\%$ on three datasets. 

\begin{table}[!t]
\renewcommand{\arraystretch}{1.1}
\caption{Accuracy ($\%$) of using different partitions of quota in the two-stage training process.}
\label{tab:quota}
\begin{center}
\setlength{\tabcolsep}{0.4cm}
\begin{tabular}{ccc}
\hline
quota split (stage$1$/stage$2$) & CIFAR100 & Mini-ImageNet \\
\hline
$10\mathrm{K}$/$37\mathrm{K}$ & $73.36$    & $45.30$         \\
$20\mathrm{K}$/$27\mathrm{K}$ & $74.10$    & $45.45$         \\
$27\mathrm{K}$/$20\mathrm{K}$ & $73.76$    & $45.54$         \\
$37\mathrm{K}$/$10\mathrm{K}$ & $73.33$    & $44.15$         \\
\hline
\end{tabular}
\end{center}
\end{table} 

In the following, we investigate (i) the split of quota between two stages and (ii) using more training stages. For (i), Table~\ref{tab:quota} shows four options of assigning $47\mathrm{K}$ quota to two stages. One can observe that the accuracy drops consistently when the first stage uses either too many or too few quotas. Intuitively, both cases will push the training paradigm towards one-stage training which is less efficient in one-bit supervision. This shows the importance of making a balanced schedule. For (ii), a three-stage training is performed on CIFAR100. We split the quota uniformly into three stages which follow the conclusions of (i), and each stage has $15\mathrm{K}$, $17\mathrm{K}$, and $15\mathrm{K}$ guesses, from the first to the last, respectively. The final test accuracy is $74.72\%$, comparable to $73.76\%$ of two-stage training. The accuracy gain brought by three-stage training is around $1\%$, which is considerable but much smaller than $2.63\%$ (two-stage training over one-stage training). Considering the tradeoff between accuracy and computational costs, we use two-stage training with an appropriate quota over two stages. 

\begin{table}[!t]
\renewcommand{\arraystretch}{1.2}
\caption{Accuracy ($\%$) of using different numbers of labeled samples for the three datasets.}
\label{tab:initial_labels}
\begin{center}
\setlength{\tabcolsep}{0.4cm}
\begin{tabular}{lccc}
\hline
\# labels per class in $\mathcal{D}^\mathrm{S}$ & $10$ & $30$ & $50$ \\
\hline
CIFAR100      & $65.06$ & $73.76$ & $73.90$ \\
Mini-ImageNet & $34.85$ & $45.54$ & $45.64$ \\
ImageNet      & $55.42$ & $60.40$ & $61.03$ \\
\hline
\end{tabular}
\end{center}
\end{table}

\subsubsection{Analysis on the Size of $\mathcal{D}^\mathrm{S}$}

We study the impact of different sizes of $\mathcal{D}^\mathrm{S}$, \textit{i.e.}, the number of full-bit labels used in the initial. Experiments are conducted on CIFAR100 and Mini-ImageNet, and the size is adjusted to $1\mathrm{K}$, $3\mathrm{K}$ (as in main experiments), $5\mathrm{K}$, and $10\mathrm{K}$ (the baseline setting without one-bit supervision); on ImageNet, the corresponding numbers are $10\mathrm{K}$, $30\mathrm{K}$ (as in main experiments), $50\mathrm{K}$, and $128\mathrm{K}$ (the baseline setting), respectively. The results in Table~\ref{tab:initial_labels} show that it is best to keep a proper amount (\textit{e.g.}, $30\%$--$50\%$) of initial labels and exchange the remaining quota to one-bit supervision. When the part of full-bit supervision is too small, the initial model may be too weak to guess enough positive labels; when it is too large, the advantage of one-bit supervision becomes small and the algorithm degenerates into a regular semi-supervised learning process. This conclusion reveals that the optimal solution is to make a balanced schedule for using supervision, including assigning the quota between the same or different types of supervision forms.

\subsubsection{Analysis on Strategies of Sampling}

We investigate the strategies of selecting samples for querying by doing experiments based on a two-stage training framework. 
By connecting to active learning, and measuring the difficulty of each sample using the top-ranked score after softmax, we investigate two sampling strategies, namely easy sampling (with the highest scores) and hard sampling (with the lowest scores). Notably, the guessing accuracy can be impacted heavily by both strategies, \textit{e.g.}, on CIFAR100, comparing to $25.3\mathrm{K}$ correct guesses obtained by random sampling, easy selection, and hard selection lead to different numbers of $30.9\mathrm{K}$ and $18.0\mathrm{K}$. The final accuracy is slightly changed from $73.76\%$ to $74.23\%$ and $74.96\%$ respectively. However, on Mini-ImageNet, the same operation causes the accuracy to drop from $45.54\%$ to $44.22\%$ and $42.57\%$ respectively. That being said, the amounts of positive labels produced by easy selection mostly are easy samples, and they can't deliver much knowledge to the model; meanwhile, the hard selection strategy mines informative labels, but produces fewer positive labels. Hence, hard selection can benefit training when the dataset is relatively easy or the initial model is strong enough, \textit{e.g.}, the accuracy is boosted by over $1\%$ on CIFAR100. However, when the initial model is not strong enough on the dataset, this strategy can harm the training, \textit{e.g.}, on Mini-ImageNet, the accuracy drops about $3\%$. Hence, we introduce self-supervised learning to enhance the initial model, to achieve success in the combination of one-bit supervision and hard sample mining. 

In summary, in the setting of one-bit supervision, developing an efficient sampling strategy is important since maximal information can be extracted from the fixed quota of querying. Some heuristic strategies are presented including using multiple stages, and performing uniform sampling. Though favorite performance is achieved, we believe that more efficient strategies still exist. 

\subsection{Results with Strategies of Class Balance}
\label{experiments:class_balance}

\begin{figure*}[htbp]
\centering
\subfloat[CIFAR100]{
\includegraphics[width=3.7cm]{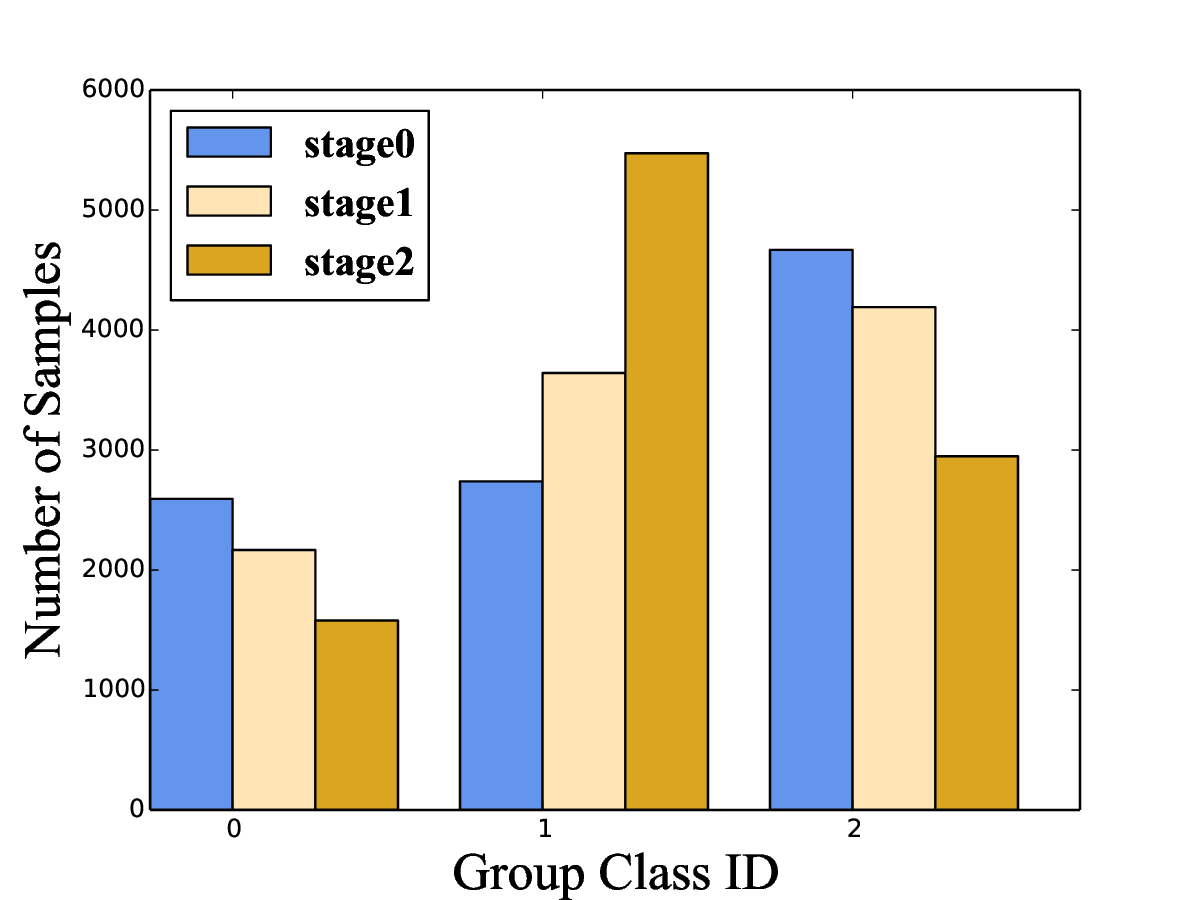}
}
\subfloat[Mini-ImageNet]{
\includegraphics[width=3.7cm]{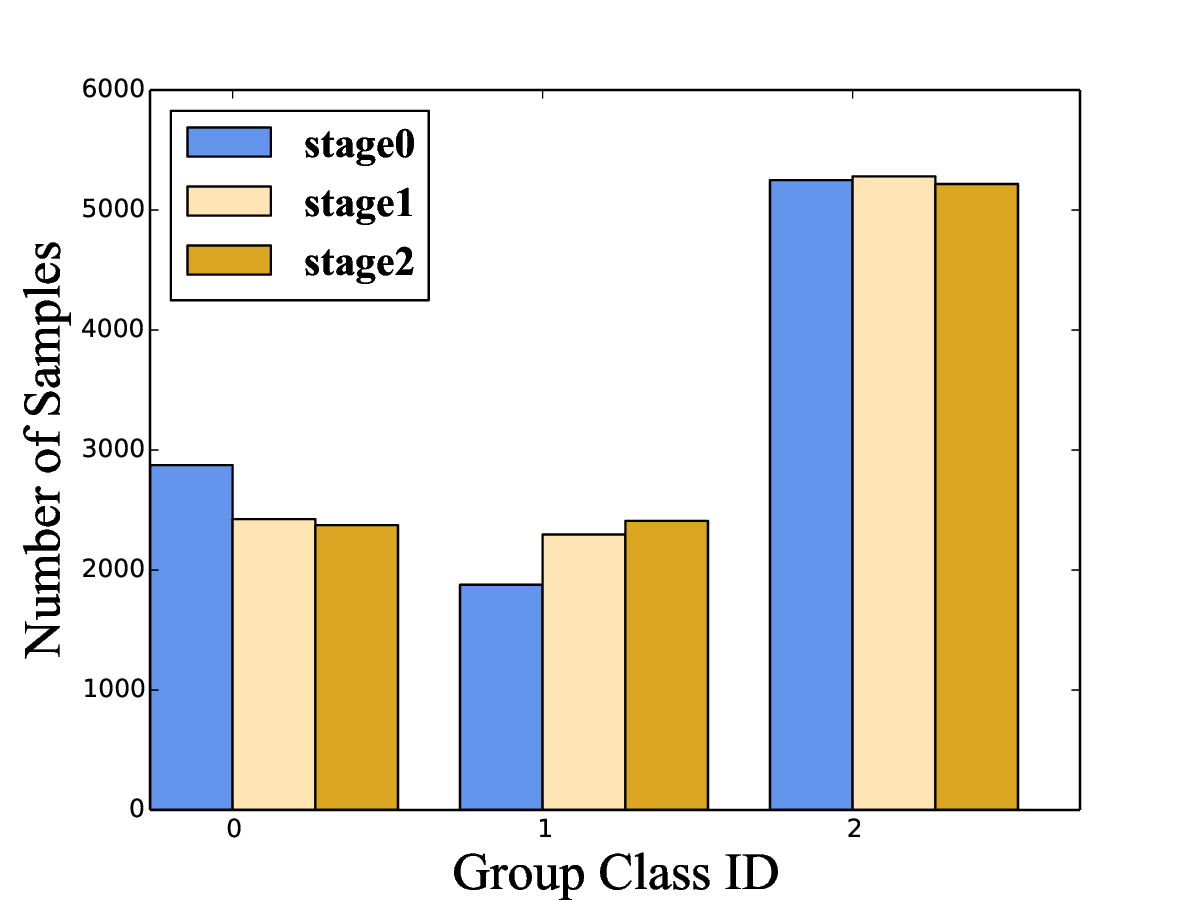}
}
\subfloat[ImageNet]{
\includegraphics[width=3.7cm]{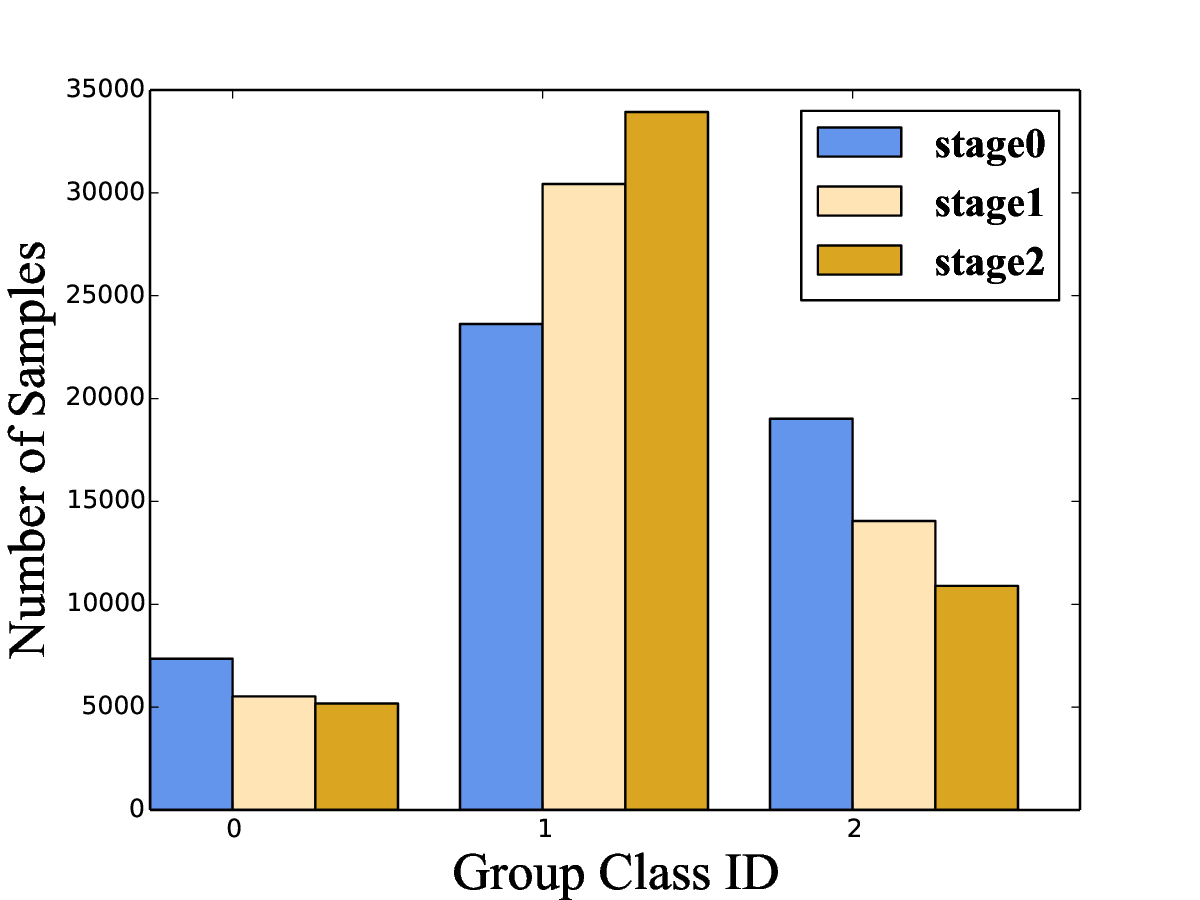}
}
\subfloat[ImageNet w/ pre-training]{
\includegraphics[width=3.7cm]{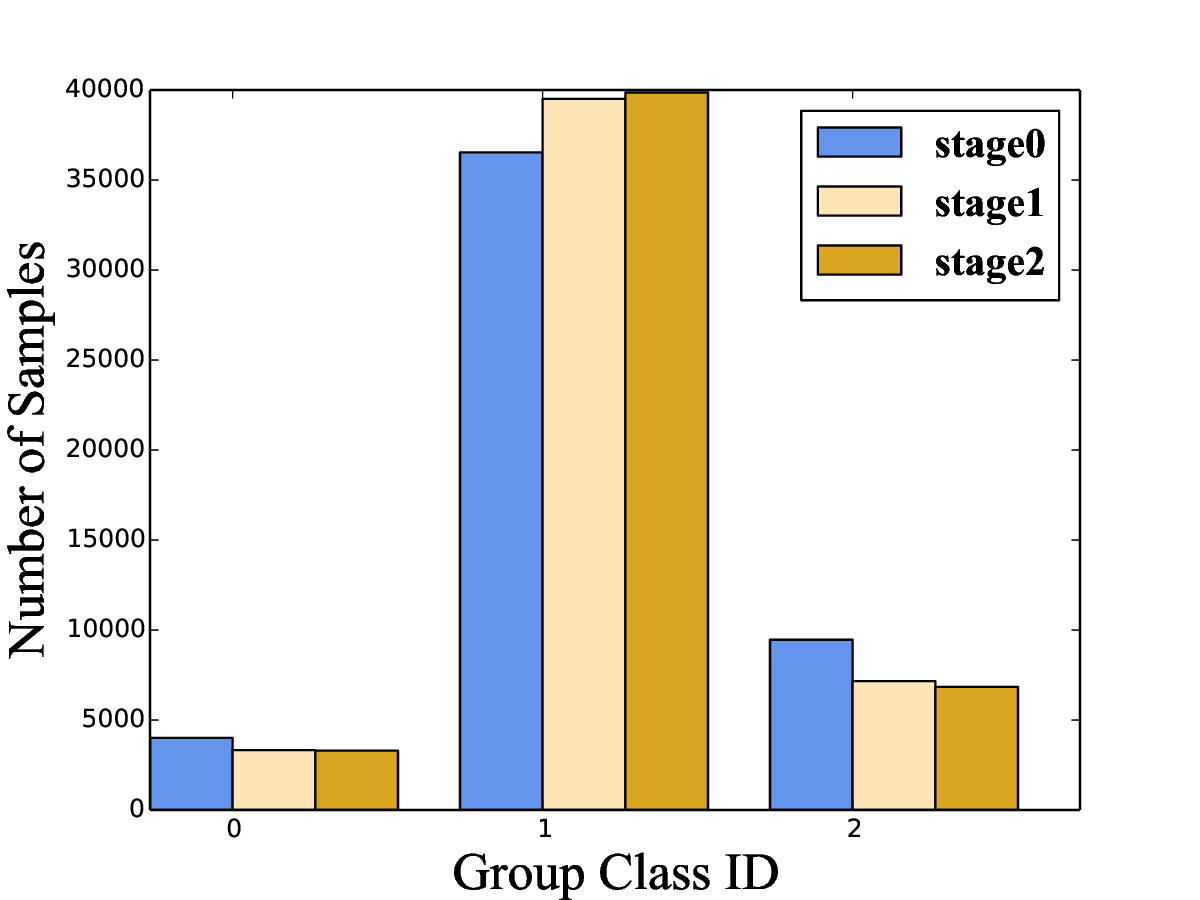}
}
\caption{The class distribution in three training stages on CIFAR100, Mini-ImageNet, and ImageNet, including that using unsupervised pre-training on ImageNet. The results are obtained by using the trained model of each stage to predict labels for the validation set. We maintain three groups to category the classes, namely group $0$ (the summation of number of samples in which class that less than $\mu\!-\!10$), group $1$ (the numbers that between $\mu\!-\!10$ and$ \mu\!+\!10$) and group $2$ (the numbers that more than $\mu\!+\!10$), where $\mu=N_{val}/C$ and $N_{val}$ is the number of samples in validation set.}
\label{fig:cb}
\end{figure*}

In the following, we investigate the developed class balance strategies. Since one-bit annotation is based on model predictions, the pseudo labels can bring class-imbalance issues and this may have a negative impact on training. We do some experiments to reveal this issue in training and the results are shown in Figure~\ref{fig:cb}. Here we summarize the number of samples in each predicted class and classify them into three groups. Ideally, every class should have around the average number of samples (corresponding to group $1$), and class imbalance happens if it is less or more than that number (corresponding to group $0$ and group $2$). From Figure~\ref{fig:cb} we can observe that this issue is obvious at stage $0$ on all three datasets. This reflects the necessity of taking strategies to relieve the issue of class imbalance, as done in our approach. As a result, the number of samples in group $1$ increases stage by stage, \textit{i.e.}, and the class imbalance is eased with the training. We also compare the results of training with/without CB on ImageNet, and the number of samples in group $1$ is $33\rm{,}930$ and $24\rm{,}678$ respectively for stage $2$, which demonstrate its effectiveness. For the model performance, as shown in the last row of Table~\ref{tab:comparison}, the proposed strategies improve the basic framework of one-bit supervision by $0.13\%$, $1.28\%$, and $4.70\%$ accuracy gains respectively on CIFAR100, Mini-ImageNet, and ImageNet, when compared to the two-stage framework with NLS. 

In addition, we notice that the gain on ImageNet is obviously larger than that on CIFAR100 and Mini-ImageNet. This is attributed to that class imbalance has a greater impact on the large-scale datasets, because (i) it has more classes ($1\rm{,}000$ vs $100$) and (ii) it has more candidate images to be annotated ($1\rm{,}276\mathrm{K}$ vs $66.9\mathrm{K}$ bits). The first means that it is more difficult for the model to predict evenly for all classes. The second means that the large number of samples can aggravate class imbalance. Hence, taking strategies to alleviate this issue on ImageNet brings more improvement. 

\subsection{Results with Self-Supervised Pre-Training}
\label{experiments:pre-train}

Here we show the experimental results of the framework incorporating SSL with one-bit supervision, which refers to section~\ref{beyond:self_supervised}. The results of using unsupervised pre-training on ImageNet are listed in Table~\ref{tab:pre-train}. Firstly we compare the results of training with and without pre-training. One can observe that, $15.14\%$ accuracy gain, a remarkable improvement is brought by incorporating self-supervised learning with one-bit supervision, which shows the potential of this new training framework. The self-supervised algorithm benefits the initial model largely so that more correct guesses are obtained, and then the model of the next stage can be strengthened further. 
Compared to the baseline which uses $10\%$ labels to fine-tune, our approach achieves $1.49\%$ accuracy gain under the same bits of supervision (following the setting in Table~\ref{tab:settings}). This further verifies the superiority of one-bit supervision when combined with self-supervised learning.

\begin{figure}[!t]
\centering
\includegraphics[width=0.7 \linewidth]{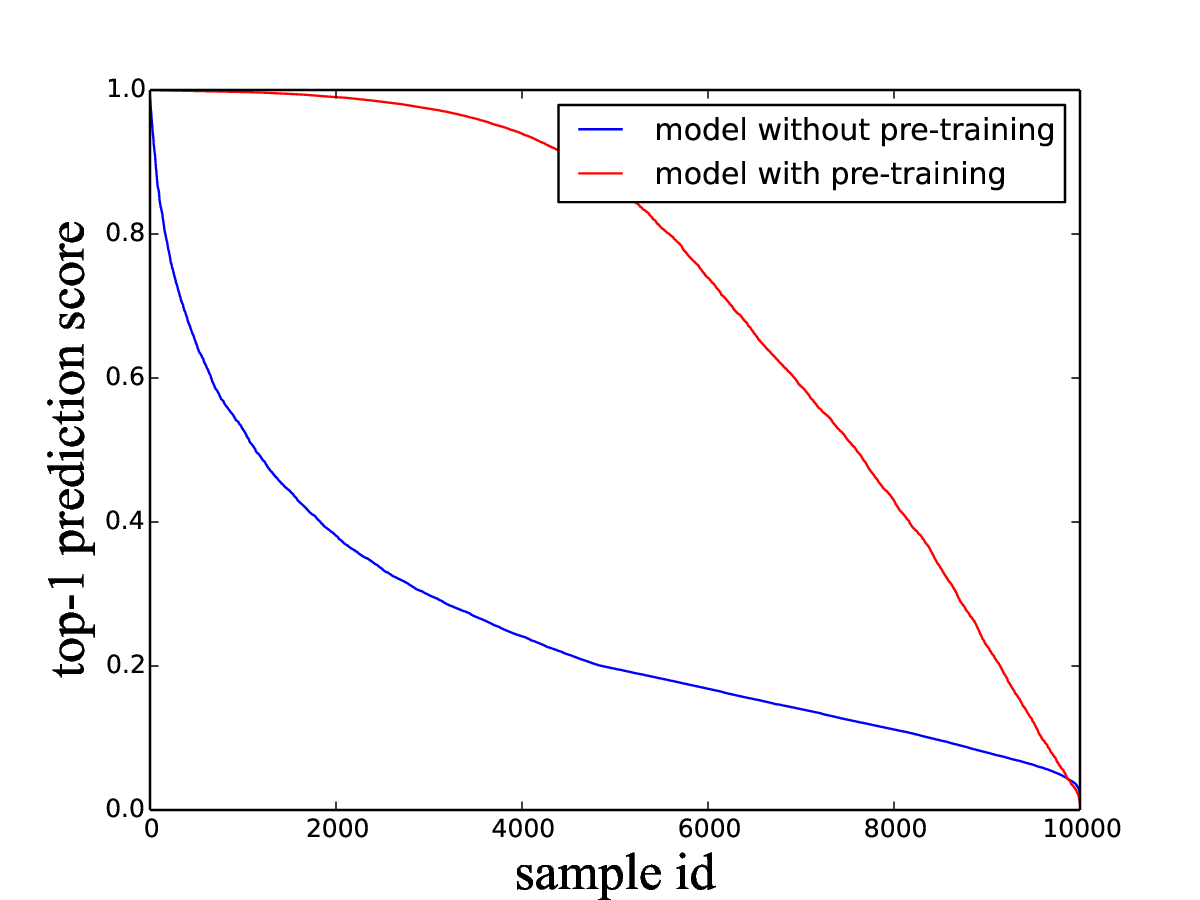}
\caption{The comparison between the top-$1$ prediction scores of the hard examples on ImageNet which are made by models with/without pre-training. The used $10\rm{,}000$ samples are selected randomly from the hard samples mined by the model without pre-training. }
\label{fig:loss_score}
\end{figure}

\begin{figure}[!t]
\centering
\includegraphics[width=0.9\linewidth]{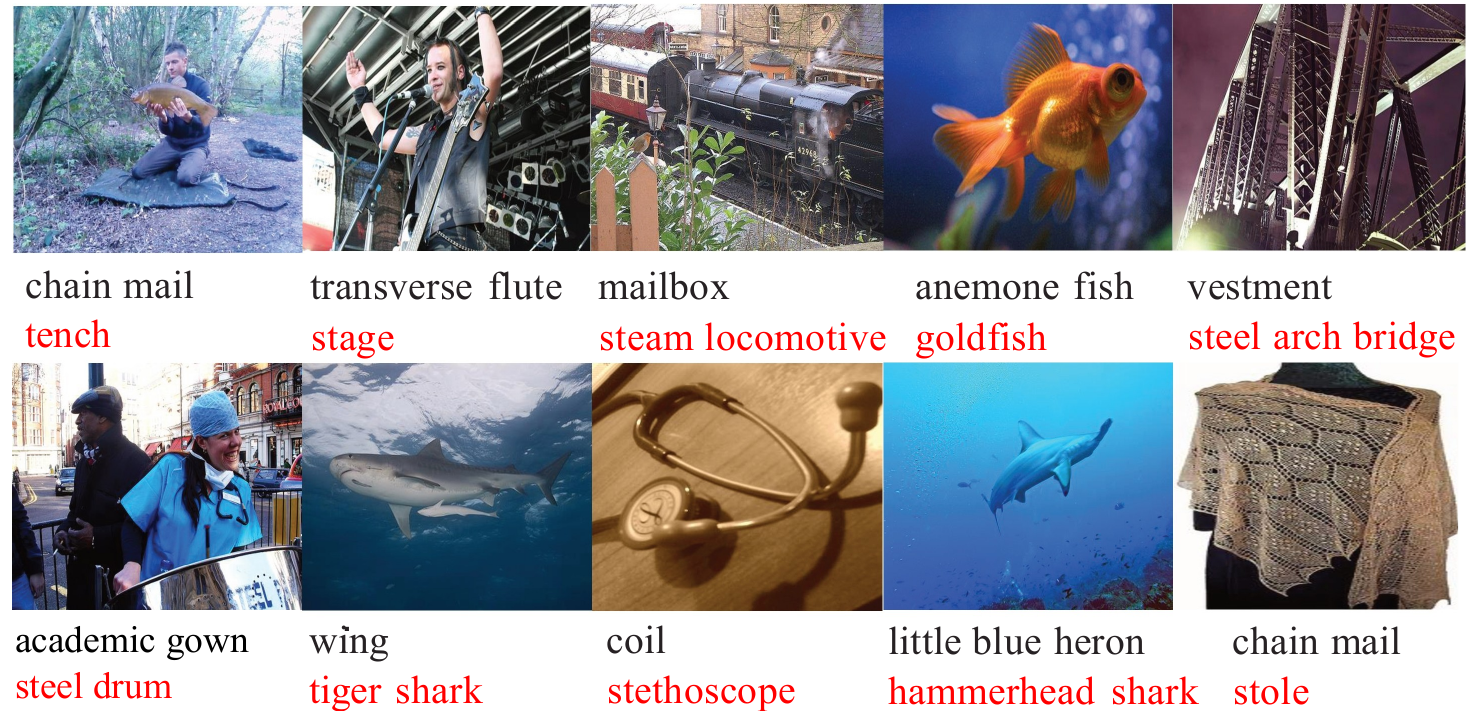}
\caption{The selected hard examples of ImageNet and their predicted labels made by the model training with/without unsupervised pre-training. The first line of the text is the predicted labels of the model without pre-training, and the second line is that of the model with pre-training. For each image, the red text indicates the correctly guessed label.}
\label{fig:hard_examples}
\end{figure}

Secondly, we investigate the impact of hard sampling (HS). From Table~\ref{tab:pre-train} we can observe that, using HS brings $4.43\%$ accuracy gain for our approach without pre-training. Though HS brings gains for the experiment without pre-training, we argue that it can not be always effective, as the discussions in Section~\ref{experiments:stages}. Hence, if the initial model is too weak to acquire enough positive labels, HS can fail. However, introducing unsupervised pre-training can address this problem by assisting in building a strong initial model. Just as the results show, using HS achieves $75.71\%$ accuracy, bringing $0.17\%$ gains for the experiments with pre-training. Considering it is based on such a high baseline, this is a significant improvement. Figure~\ref{fig:loss_score} shows the top-$1$ prediction scores ($p_c$) on the selected hard samples made by the model with and without pre-training. It reveals that the model with pre-training obtains higher scores than that without pre-training in most cases. According to the deduction in Section~\ref{approach:mathematical}, this demonstrates that using unsupervised pre-training makes it be successful to mine hard samples for one-bit supervision. Figure~\ref{fig:hard_examples} presents some mined hard examples. We can observe that many incorrect guesses made by the model without pre-training can be corrected by using it. And, this shows the potential of using active learning to improve one-bit supervision by annotating informative samples. 

Thirdly we investigate the strategies of class balance for this new framework. As the discussions in Section~\ref{experiments:class_balance}, alleviating the issue of class imbalance is important for datasets with more classes. The subfigure~(d) in Figure~\ref{fig:cb} shows that class imbalance still exists in the model initialized with pre-trained weights. The results in Table~\ref{tab:pre-train} show that using CB brings $0.66\%$ accuracy gain. Both the gain achieved and the distribution improvement in subfigure~(d) verify that CB is effective in relieving class imbalance issues for this framework. Finally, we underline that this new framework achieves $76.37\%$ top-$1$ accuracy, which is superior to many state-of-the-art semi-supervised approaches, and is comparable to the result of a standard supervised ResNet-$50$ trained using all labels ($76.5\%$). This shows the success of the developed framework which combines self-supervised learning with one-bit supervision. 

\begin{table}[!t]
\renewcommand{\arraystretch}{1.4}
\caption{The results for experiments using unsupervised pre-training on ImageNet, which corresponds to the extended framework with SSL. The upper lists two SOTA semi-supervised methods UDA and FixMatch (using $10\%$ labels). and the unsupervised pre-trained model HSA (fine-tune on $10\%$ labels). The middle shows the results of our approach without unsupervised pre-training. The below shows the results using unsupervised pre-training. Here, PT, HS, and CB indicate pre-training, hard sampling, and class balancing, respectively.}
\label{tab:pre-train}
\begin{center}
\setlength{\tabcolsep}{0.4cm}
\begin{tabular}{lcccc}
\hline
Method                    & PT & HS & CB & ImageNet \\ \hline
UDA~\cite{xie2019unsupervised} & -- & -- & -- & $68.80$ \\
FixMatch~\cite{sohn2020fixmatch} & -- & -- & -- & $71.50$ \\
HSA~\cite{xu2020hierarchical}    & -- & -- & -- & $74.05$ \\ \hline
Ours & & & &$47.83\rightarrow57.44\rightarrow60.40$\\ 
Ours & & \checkmark & & $47.83\rightarrow59.33\rightarrow64.83$ \\ 
Ours & & & \checkmark & $47.83\rightarrow59.41\rightarrow65.10$ \\  \hline
Ours & \checkmark & & & $70.51\rightarrow73.95\rightarrow75.54$ \\ 
Ours & \checkmark & \checkmark & & $70.51\rightarrow74.33\rightarrow75.71$ \\ 
Ours & \checkmark & \checkmark & \checkmark & $70.51\rightarrow75.31\rightarrow76.37$ \\ \hline
\end{tabular}
\end{center}
\end{table}

\subsection{Results with Unsupervised Domain Adaptation}
\label{experiments:transfer}

The performance of combining UDA with one-bit supervision which refers to section~\ref{beyond:transfer} is shown in this part. We evaluate this new framework which trains without using initial full-bit labels on DomainNet, and two adaptation experiments are conducted, namely, Clip to Real and Quickdraw to Real. The results are listed in Table~\ref{tab:transfer}. We can observe that, for Clip$\rightarrow$Real, the two-stage training of one-bit supervision without using source data achieves $83.70\%$ accuracy, by using $18.22\%$ of supervision and achieving $99.48\%$ of performance of full-supervised training ($84.14\%$). When combined with active learning the accuracy becomes $81.57\%$, which means using only $11.86\%$ of supervision and achieving $96.95\%$ of performance of full-supervised training. These results reveal the superiority of one-bit supervision on the efficient utilization of supervision information. Also, the framework of pure one-bit supervision via UDA enlarges the advantage of one-bit annotation. In addition, we also observe that using source data brings a negative impact in the latter training stages, \textit{e.g.}, it achieves $77.05\%$ accuracy when using the same amount of supervision with that using no source data. We attribute it to that one-bit annotation brings a certain amount of supervision to the target domain, and still using source data can degrade its generalization performance. 

For the experiments Quickdraw$\rightarrow$Real, a much harder transfer task that provides a weaker initial model, we conduct a three-stage training framework. Basically, similar observations can be obtained from these results. Firstly, training without source data achieves $83.48\%$ accuracy, $99.22\%$ of performance of the full-supervised training, by using $23.34\%$ of supervision. The result of incorporating active learning is $82.37\%$, which is also very close to the full-supervised results and it uses $17.79\%$ of supervision. Also, using source data to train achieves a lower accuracy ($65.68\%$) when using the same bits of supervision. These two experiments on DomainNet verify the advantage of the proposed framework on annotation saving. 

\begin{table*}[]
\setlength{\tabcolsep}{0.4cm}
\renewcommand{\arraystretch}{1.4}
\caption{The results for experiments that involve UDA on DomainNet, which corresponds to the extended framework with UDA. Besides the upper bound of full-supervised training using all labels, we list the results of three methods which respectively are one-bit supervision with source data, without source data, and with active learning, as well as the bits of supervision they used. These experiments are conducted to verify the superiority of our approach in annotation saving. }
\label{tab:transfer}
\begin{center}
\begin{tabular}{lccc}
\hline
Methods             & Dataset         & $\#$ of Supervision  & Accuracy (\%)                   \\ \hline
Full-supervised     & Real            & $100\%$                  & $84.14$               \\ \hline
Ours (w/source)    & \multirow{3}{*}{Clip $\rightarrow$ Real}    & $18.22\%$  & $54.53\rightarrow70.69\rightarrow77.05$    \\
Ours (wo/source)   &                                       & $18.22\%$    & $54.53\rightarrow77.13\rightarrow83.70$      \\
Ours + AL          &                                       & $11.86\% $   & $54.53\rightarrow65.26\rightarrow81.57$  \\ \hline
Ours (w/source)  & \multirow{3}{*}{Quickdraw $\rightarrow$ Real} & $23.34\%$   & $19.06\rightarrow52.62\rightarrow61.99\rightarrow65.68$ \\
Ours (wo/source) &                      & $23.34\%$    & $19.06\rightarrow70.39\rightarrow83.18\rightarrow83.48$ \\
Ours + AL        &                      & $17.79\%$    & $19.06\rightarrow65.59\rightarrow80.02\rightarrow82.37$ \\ \hline
\end{tabular}
\end{center}
\end{table*}

\section{Conclusion and Future Works}
\label{conclusions}

In this paper, we propose a new learning methodology named one-bit supervision. Compared to conventional approaches which need to annotate the correct label for each image, our system annotates an image by answering a yes-or-no question for the guess made by a trained model. A multi-stage training framework is designed to acquire more correct guesses. Meanwhile, we propose a method of negative label suppression to utilize incorrect guesses. We provide mathematical foundations for the proposed approach from three aspects, namely one-bit annotation is more efficient than full-bit annotation in most cases, the solution to one-bit supervision is to query by the class with the largest predicted probability, and when the predicted probability is large enough mining hard examples can improve our approach. Experiments on three popular benchmarks verify that the basic framework outperforms semi-supervised learning under the same bits of supervision. To extend the basic approach, we design two new frameworks by incorporating it with self-supervised learning and unsupervised domain adaptation. The first benefits from active learning and achieves remarkable performance on ImageNet. Also, two strategies are used to alleviate class imbalance in training. The second conducts pure one-bit annotation on the target dataset and enjoys the superiority in annotations saving, which is evaluated on DomainNet. 


One-bit supervision is a new learning paradigm that leaves a few open problems. For example, we have investigated one-bit supervision in image classification, and it is interesting to extend this framework to other vision tasks such as object detection and semantic segmentation. This is related to some prior efforts such as~\cite{papadopoulos2017training,xu2016deep}. For detection, the cost will be much lower if the labeler just needs to annotate whether a detected bounding box is correct (\textit{e.g.}, has an IOU no lower than a given threshold to a ground-truth object); for segmentation, the labeler is given the segmentation map and determines whether each instance or a background region has a satisfying IOU. These problems are also more challenging than image classification but they have higher values in real-world applications.

\begin{acks}
This work was supported by the National Key Research and Development Program of China under grant 2019YFA0706200, 2018AAA0102002, and in part by the National Natural Science Foundation of China under grant 61932009.
\end{acks}

\bibliographystyle{ACM-Reference-Format}
\bibliography{one_bit}


\end{document}